\documentclass{article}
\usepackage{spconf,amsmath,graphicx}

\usepackage{booktabs}
\usepackage{times}
\usepackage{latexsym}
\usepackage{url}
\usepackage{amssymb}
\usepackage{bm}
\usepackage{multirow}
\usepackage{enumitem}
\newcommand{\vect}[1]{\bm{#1}}
\newcommand{\mat}[1]{\bf{#1}}

\usepackage{color}

\title{Transfer Learning for Context-Aware Spoken Language Understanding}
\name{Qian Chen, Zhu Zhuo, Wen Wang, Qiuyun Xu}
\address{Speech Lab, DAMO Academy, Alibaba Group\\
\small\texttt{\{tanqing.cq,~zhuozhu.zz,~w.wang,~qiuyun.xqy\}@alibaba-inc.com}
}
\begin{document}
%\ninept
%
\maketitle
\begin{abstract}
Spoken language understanding (SLU) is a key component of task-oriented dialogue systems. SLU parses natural language user utterances into semantic frames. Previous work has shown that incorporating context information significantly improves SLU performance for multi-turn dialogues. However, collecting a large-scale human-labeled multi-turn dialogue corpus for the target domains is complex and costly. To reduce dependency on the collection and annotation effort, we propose a Context Encoding Language Transformer (CELT) model facilitating exploiting various context information for SLU. We explore different transfer learning approaches to reduce dependency on data collection and annotation. In addition to unsupervised pre-training using large-scale general purpose unlabeled corpora, such as Wikipedia, we explore unsupervised and supervised adaptive training approaches for transfer learning to benefit from other in-domain and out-of-domain dialogue corpora. Experimental results demonstrate that the proposed model with the proposed transfer learning approaches achieves significant improvement on the SLU performance over state-of-the-art models on two large-scale single-turn dialogue benchmarks and one large-scale multi-turn dialogue benchmark.
\end{abstract}
\begin{keywords}
Transfer Learning, Spoken Language Understanding, Transformer
\end{keywords}

\section{Introduction}
\begin{figure}[htb]
\centering
\includegraphics[width=0.3\textwidth]{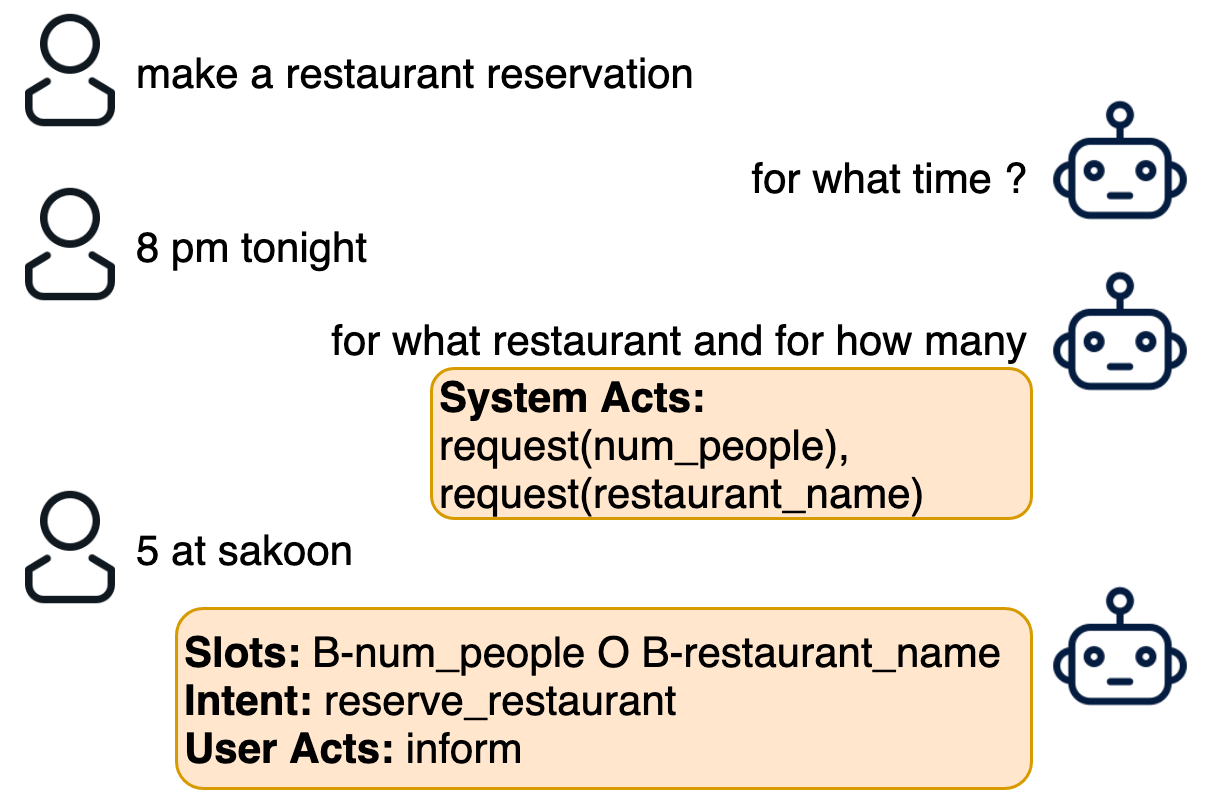}
\caption{An example user query in a multi-turn conversation and its semantic frame with slot, intent and user dialogue act annotations. For user query ``5 at sakoon'', ``5'' could indicate date, time, number of people, etc; yet with the context, it is most likely resolved as the number of people. }
\label{fig:example}
\end{figure}

Spoken language understanding (SLU) is a key component of task-oriented dialogue systems, which assist user to complete tasks such as  booking flight tickets. SLU parses user utterances into semantic frames, including intents, slots, and user dialogue acts~\cite{tur2011spoken}. The semantic frame for a restaurant reservation query is shown in Figure~\ref{fig:example}. Both intents and user dialogue acts represent the user's intentions, but intents and user acts could be defined with different granularities. In this work, we model both intent and user act classification when both of them are available in the dialogue corpora.

Previous research in SLU has significantly focused on single-turn SLU, that is, understanding the current user utterance. However, completing a task usually necessitates multiple turns of back-and-forth conversations between the user and the system. Multi-turn SLU imposes different challenges from single-turn SLU, for example, entities introduced earlier in conversation may be referred later by the user and the system, information mentioned earlier may be skipped later, causing ambiguities, as shown in Figure~\ref{fig:example}. Incorporating contextual information has been shown useful for multi-turn SLU~\cite{DBLP:conf/icassp/BhargavaCHS13,DBLP:conf/icassp/XuS14,DBLP:conf/icmi/ChenSRG15,DBLP:conf/iui/SunCR16,DBLP:conf/icassp/ShiYCPHP15,DBLP:conf/interspeech/GuptaRH18}. Information from previous intra-session utterances was explored by applying SVM-HMMs to sequence tagging for SLU~\cite{DBLP:conf/icassp/BhargavaCHS13}. Contextual information was incorporated into the recurrent neural network (RNN) structure~\cite{DBLP:conf/icassp/XuS14,DBLP:conf/icassp/ShiYCPHP15}. Chen et al.~\cite{DBLP:conf/interspeech/ChenHTGD16} proposed a memory network based approach for multi-turn SLU by encoding history utterances and leveraging the memory embeddings through attention. Bapna et al.~\cite{DBLP:conf/sigdial/BapnaTHH17} enhanced the memory network architecture by adding a BiRNN session encoder temporally combining the current utterance encoding and the memory vectors. Su et al.~\cite{DBLP:conf/naacl/SuYC18} investigated different time-decay attention mechanisms. Gupta el al.~\cite{DBLP:conf/interspeech/GuptaRH18} proposed an approach to encode system dialogue acts for SLU, substituting the use of system utterances. Also, various models have been proposed for jointly modeling intent and slot predictions and achieved significant performance improvement over models that model these predictions independently~\cite{DBLP:conf/asru/XuS13,DBLP:conf/interspeech/Hakkani-TurTCCG16,DBLP:conf/ijcai/ZhangW16a,DBLP:conf/interspeech/LiuL16,DBLP:conf/naacl/GooGHHCHC18,DBLP:journals/corr/abs-1812-09471}. In this work, we also follow the joint learning paradigm.

However, lack of human-labeled data for SLU 
%and other natural language processing (NLP) tasks 
results in poor generalization capability.
A variety of transfer learning (TL) techniques were proposed for addressing the data sparsity challenge. One category of TL approaches includes training general purpose language representation models using a large amount of unlabeled text, such as ELMo~\cite{DBLP:conf/naacl/PetersNIGCLZ18}, GPT~\cite{DBLP:techreport/ge1ne8r}, and BERT~\cite{DBLP:journals/corr/abs-1810-04805}. Pre-trained models can be fine-tuned on NLP tasks and have achieved significant improvement over training on the task-specific annotated data. Bapna et al.~\cite{DBLP:journals/corr/BapnaTHH17aa} leveraged slot name and description encodings within a multi-task model for domain adaptation. Lee et al.~\cite{DBLP:journals/aaai/Lee19} proposed zero-shot adaptive transfer for slot tagging by embedding the slot descriptions and fine-tuning a pre-trained model on the target domain.  Siddhant et al.~\cite{DBLP:journals/aaai/Siddhant19} used a light-weight ELMo model for pre-training and unsupervised and supervised transfer.

Our contribution in this paper is threefold: \textbf{First}, we propose a Context Encoding Language Transformer (CELT) model for context-aware SLU. Different from previous work of exploring various encoding schemes and attention mechanisms to encode context for multi-turn SLU, CELT facilitates encoding various context information in the dialogue history for SLU, such as user and system utterances, speaker information, system acts, and utilizing these information in a unified framework through a multi-head self-attention mechanism. The context information that CELT can exploit is extensible. For example, for a conversational system that facilitates the use of a screen for multi-modal interactions, screen-displayed information can be treated similarly as context in CELT and help understand user query. \textbf{Second}, we develop a multi-step TL approach on CELT, namely, unsupervised pre-training to exploit large-scale general purpose unlabeled text, unsupervised adaptive training and supervised adaptive training to exploit other in-domain and out-of-domain dialogue corpora. To our knowledge, the CELT model and the multi-step TL approach on CELT are first proposed in this work for multi-turn SLU.  \textbf{Third}, we systematically evaluate the efficacy on SLU from various context information and TL approaches. The proposed CELT model together with the proposed TL approaches significantly outperform the state-of-the-art performance on two large-scale single-turn dialogue benchmarks and one large-scale multi-turn dialogue benchmark.

\section{Proposed Approach}
\begin{figure}[t]
\centering
\includegraphics[width=0.45\textwidth]{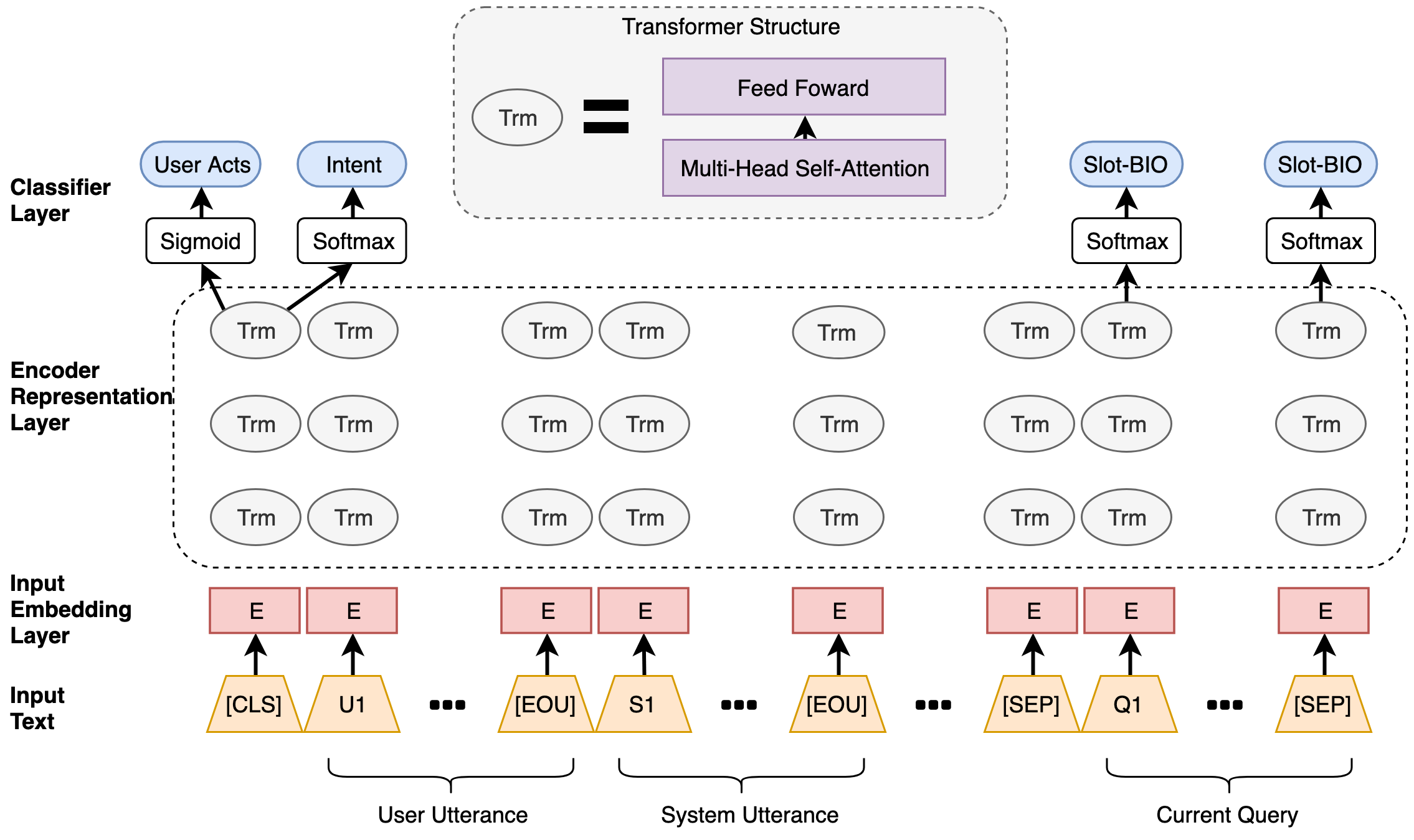}
\caption{A high-level view of the CELT model. It consists of the input embedding layer, the encoder representation layer, and the final classifier layer. Details of the input embedding layer are illustrated in Figure~\ref{fig:input_embedding}. ``Trm'' denotes Transformer blocks. }
\label{fig:model}
\end{figure}

Figure~\ref{fig:model} provides a high-level illustration of CELT, which consists of the input embedding layer, the encoder representation layer, and the final classifier layer.

\subsection{Input Embedding Layer}

Given the current query token sequence ${\vect q=q_1,\dots, q_T}$ 
at turn ${t}$, and previous turns in the dialogue session, i.e., user turns ${\vect u}^i, i=[1, \dots, t-1]$, and system turns ${\vect s}^i, i=[1, \dots, t-1]$, the target is to predict the semantic frame, including intent, user acts, and slots, for ${\vect q}$. We concatenate all the previous turns chronologically in a dialogue session and the current query as the input text ${\vect x}$, i.e., ${\vect x} = ({\vect u}^1,{\vect s}^1,\dots, {\vect u}^{t-1}, {\vect s}^{t-1}, {\vect q})$. The first token of every input text is always the special classification embedding ([CLS]) which is used to predict the intent and user acts. Each utterance in the previous turns is inserted an end-of-utterance ([EOU]) token. The previous utterances and the current turn are separated by a special token ([SEP]).

\begin{figure}[t]
\centering
\includegraphics[width=0.45\textwidth]{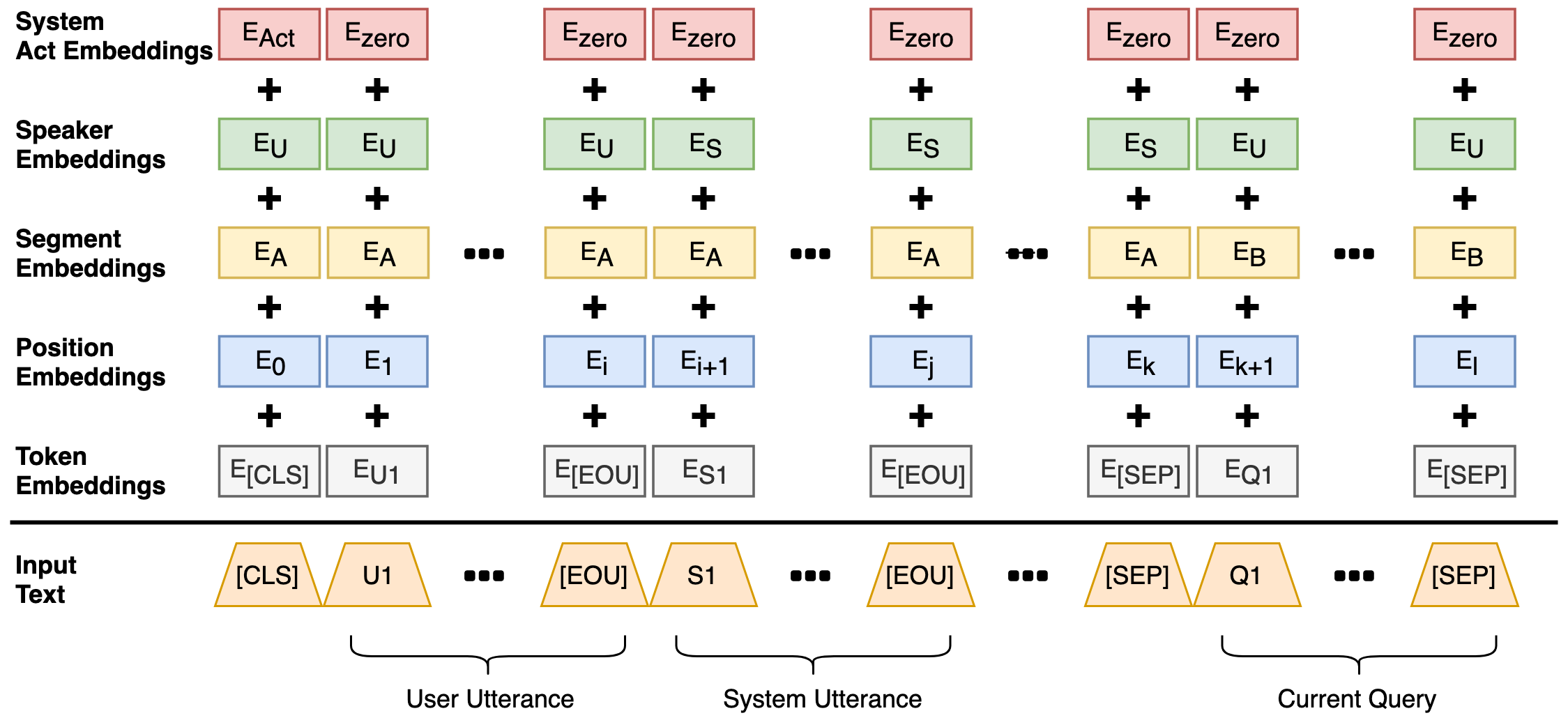}
\caption{The input embedding layer of CELT.}
\label{fig:input_embedding}
\end{figure}

For a token in the input text, its input embedding is an element-wise sum of token embeddings, position embeddings, segment embeddings, and embeddings of other context information. In this work, we add speaker embeddings and system act embeddings into the sum to obtain the final input embedding. Figure~\ref{fig:input_embedding} illustrates the input embedding layer.

The learned \textit{WordPiece embeddings}~\cite{DBLP:journals/corr/WuSCLNMKCGMKSJL16} are used to alleviate the out-of-vocabulary (OOV) problem. The learned \textit{position embeddings} are used to capture the sequence order information. The learned \textit{segment embeddings} are used to distinguish the previous turns and the current query, hence all previous turns have the same segment embeddings. \textit{Speaker embeddings} are used to distinguish the user's turns or the system's turns, considering that speaker role information has been shown useful for SLU in complex dialogues~\cite{DBLP:conf/naacl/SuYC18}. \textit{System act embeddings} encode the system act information. Each system act contains an act type and optional slot and value parameters. The acts are categorized into two broad types: acts with an associated slot (i.e. request(date), select(time=7pm)) 
and acts without associated slots (e.g. negate). We keep the slot type, ignore the slot values, and convert the system acts into embeddings just like word embeddings. We define an n-hot binary vector ${\vect a}$ to represent system acts of the previous system turn in the system act vocabulary $A$\footnote{We only use the system acts from the previous system turn instead of using all system acts in the dialogue history, in order to keep the same setting as Gupta et al.~\cite{DBLP:conf/interspeech/GuptaRH18}.}, and convert the n-hot binary vector to a fixed-sized vector by multiplying it with a system act embedding matrix ${\mat E}^a$, that is, ${\vect e}^a = {\mat E}^a {\vect a}$.

\subsection{Encoder Representation Layer}

The encoder representation layer is a multi-layer Transformer~\cite{DBLP:conf/nips/VaswaniSPUJGKP17} consisting of multi-head self-attention sub-layer and feed-forward sub-layer in each layer. The multi-head self-attention mechanism builds upon scaled dot-product attention, operating on query $Q$, key $K$, and value $V$~\cite{DBLP:conf/nips/VaswaniSPUJGKP17}:
\begin{align}
\mathrm{Attention}(Q,K,V) = \mathrm{softmax}(\frac{Q K^T}{\sqrt{d_k}})V \,,
\label{equ:attention}
\end{align}
\noindent where $d_k$ is the dimension of the keys. Multi-head attention mechanisms obtain $h$ different representations of $(Q, K, V)$, compute scaled dot-product attention for each representation, and concatenate the results. This can be expressed in the same notation as Equation (\ref{equ:attention}):
\begin{align}
\mathrm{head}_i = \mathrm{Attention}(QW_i^Q,KW_i^K,VW_i^V) \,, \\
\mathrm{MultiHead(Q,K,V)} = \mathrm{Concat}_i(\mathrm{head}_i)W^O \,,
\end{align}
\noindent where $W_i^Q,W_i^K,W_i^V,W^O$ are projection matrices~\cite{DBLP:conf/nips/VaswaniSPUJGKP17}.

The concatenation is projected with the feed-forward neural network (FFN) sub-layer. We propose using a two-layered network with GELU~\cite{DBLP:journals/corr/HendrycksG16} activation. Given trainable weights $W_1$, $W_2$, $b_1$, $b_2$, this sub-layer is defined as:
\begin{align}
\mathrm{FFN}(x) = W_2 \mathrm{GELU}(W_1x + b_1) + b_2 \,,
\end{align}

After feeding the input embedding sequence into the encoder representation layer,
the output hidden states are ${\mat H} = ({\vect h}_1,\dots, {\vect h}_T)$, where ${\vect h}_1$ corresponds to [CLS].

\subsection{Final Classifier Layer}
\subsubsection{Intent and user act classification}
We assume that each user query contains a single intent and multiple user dialogue acts. Intent classification (IC) predicts the intent probability distribution $p^i$, using Equation~(\ref{equ:intent}). User act classification (UAC) is defined as a multi-label binary classification problem. 
The probability of the presence of the $k$-th user act in the user query, $p^a(k)$, is calculated by Equation~(\ref{equ:act}).
\begin{align}
\label{equ:intent}
p^i = \mathrm{softmax}({\mat W}_i F_i({\vect h}_{[CLS]}) + {\vect b}_i) \,, \\
\label{equ:act}
p^a = \mathrm{sigmoid}({\mat W}_a F_a({\vect h}_{[CLS]}) + {\vect b}_a) \,,
\end{align}
\noindent where $F_i$ and $F_a$ are non-linear feed-forward layers with tanh activation. During inference, the intent label is predicted by $\mathrm{argmax}(p^i)$, and user acts are predicted when the probability $p^a(k)$ is greater than the threshold $t_u$, where $t_u$ is a hyperparameter tuned on the validation set. 

\subsubsection{Slot filling}
Slot filling (SF) identifies the values for different slots present in the user utterance. We use the BIO (begin-inside-outside) tagging scheme to assign a label to each token. We feed the final hidden states $\vect{h}_2,\dots,\vect{h}_T$ into a softmax layer to classify over the SF labels. To make this procedure compatible with the WordPiece tokenization, we feed each tokenized input word into a WordPiece tokenizer and use the hidden state corresponding to the first sub-token as the input to the softmax classifier. 
\begin{align}
p^s_i = \mathrm{softmax}({\mat W}_s F_s({\vect h}_i) + {\vect b}_s) \,,\
\end{align}
\noindent where \({\vect h}_i\) is the hidden state corresponding to the first sub-token of word \(q_i\) in the current query ${\vect q}$, and $F_s$ is a non-linear feed-forward layer with GELU activation.

For joint learning, the objective is to minimize the sum of the softmax cross-entropy losses of IC and SF and the sigmoid cross-entropy loss of UAC. Previous work has shown that an additional CRF layer on top of a BiLSTM can improve the performance for sequence tagging~\cite{DBLP:conf/acl/ZhouX15,DBLP:journals/corr/HuangXY15}. Hence we investigate the efficacy of adding a CRF layer for modeling slot label dependencies, on top of the Encoder Representation Layer, similar to~\cite{DBLP:conf/icml/LaffertyMP01}.

\subsection{Transfer Learning}
To improve SLU on the target domains and reduce dependency on data collection and annotation, we explore TL based on the CELT model, by leveraging large-scale unlabeled text and other multi-turn dialogue corpora either unlabeled or labeled with different intent/slot/dialog act labels, from the same or different domains w.r.t. the target domains. We develop a multi-step transfer learning approach. In the \textit{first} step, to exploit large-scale unlabeled text, we use unsupervised pre-training based on the BERT model with two tasks trained together, i.e., masked language model (MLM) and next sentence prediction (NSP)~\cite{DBLP:journals/corr/abs-1810-04805}. The resulting model is denoted \(\theta_A\). Next, to exploit other dialogue corpora, we propose two TL methods, namely, unsupervised adaptive training and supervised adaptive training. In the \textit{second} step, the unsupervised adaptive training approach trains \(\theta_A\) on other unlabeled dialogue corpora, using the MLM and NSP losses. The resulting model is denoted \(\theta_B\). In the \textit{third} step, given other labeled dialogue corpora, the proposed supervised adaptive training approach fine-tunes \(\theta_B\) on the labeled data, based on the combined loss of IC and SF\footnote{Note that in the combined loss, when samples have multiple intent labels, sigmoid cross-entropy loss is used for IC instead of softmax cross-entropy loss.}. The resulting weights for the input embedding layer and the encoder representation layer of CELT are then used to initialize the new CELT model. This model is denoted \(\theta_C\) for the next step fine-tuning for the target domain SLU. This way, our supervised adaptive training can exploit labeled data with intent/slot labels different from labels used for the target domains. In the \textit{fourth} step, \(\theta_C\) is fine-tuned on the target domain labeled data based on the combined loss of IC, SF, and UAC.

\section{Experiments and Analysis}

\subsection{Data}
We conduct two sets of experiments. For the first set of experiments, we evaluate the efficacy of BERT pre-train and joint modeling of IC and SF in CELT on the single-turn ATIS~\cite{DBLP:conf/slt/TurHH10} and Snips~\cite{DBLP:journals/corr/abs-1805-10190} dialogue corpora. ATIS includes audio recordings of people making flight reservations. Snips is collected from the Snips personal voice assistant. We use the same data division as~\cite{DBLP:conf/naacl/GooGHHCHC18} for both datasets. The data statistics are summarized in Table~\ref{tab:statistics}. For these two datasets, system act embeddings and user act classifier are not used, because system and user dialogue acts are not annotated. Speaker embeddings are not used since there is only the current query in single-turn dialogues.

\begin{table}[ht!]
\begin{center}
\scalebox{0.9}{
\begin{tabular}{l | r r r}
\hline
\textbf{Dataset} & \textbf{Snips} &\textbf{ATIS} & \textbf{GSD} \\
\hline
Intents & 7 & 21 & 3* \\
Slots & 72 & 120 & 21 \\
User Act & -& - & 22 \\
Training samples & 13,084 & 4,478 & 8,148\\
Validation samples & 700 & 500 & 2,116 \\
Test samples & 700 & 893 & 4,800 \\
\hline
\end{tabular}
}
\end{center}
\caption{Statistics for the Snips, ATIS and GSD data sets, including the number of intent types, slot labels (after applying the BIO scheme on the original slots, including O) and user acts for the training set, the number of samples in the training, validation, and test sets, respectively. *: note that for GSD, intents are quite high-level while the user acts have the same level of granularity as intents for Snips and ATIS.}
\label{tab:statistics}
\end{table}

For the second set of experiments, we evaluate the proposed model and TL approaches on the multi-turn Google Simulated Dialogues (GSD)\footnote{https://github.com/google-research-datasets/simulated-dialogue}~\cite{DBLP:conf/interspeech/GuptaRH18}. We explore Microsoft Dialogue Challenge (MDC)\footnote{https://github.com/xiul-msr/e2e\_dialog\_challenge}~\cite{DBLP:journals/corr/abs-1807-11125} and MultiWOZ 2.0 (WOZ)\footnote{https://www.repository.cam.ac.uk/handle/1810/280608. Note that WOZ does not have user act annotations so cannot be directly used for SLU.}~\cite{DBLP:conf/emnlp/BudzianowskiWTC18} datasets as other dialogue corpora for evaluating the proposed TL approaches.
We use the same data division as~\cite{DBLP:conf/interspeech/GuptaRH18}.
The GSD dataset covers restaurant (GSD-Resturant) and movie (GSD-Moive) domains. The entire GSD dataset (GSD-Overall) consists of 3 intents, 12 slot types, and 22 user dialogue act types. The data statistics are summarized in Table~\ref{tab:statistics}. Note that the 3 intents (``BUY\_MOVIE\_TICKETS",
``FIND\_RESTAURANT", ``RESERVE\_RESTAURANT") are quite high-level; instead, the 22 user dialog act types provide the user intent information for the SLU task.
The MDC dataset covers restaurant, movie, and taxi domains, with 4103, 2890, and 3094 training dialogues, 11 intents, and 30, 29, and 19 slots for the three domains, respectively. The WOZ dataset consists of human-human written conversations spanning 7 domains and 10,438 dialogues in total. 

\subsection{Training Details}
The Transformer block in CELT has 12 layers, 768 hidden states, 3072 feed-forward size, and 12 self-attention heads. The size of hidden states in the final classifier layer is 768. 
For pre-training, we use the English uncased BERT-Base model\footnote{https://github.com/google-research/bert}, pre-trained on the BooksCorpus~\cite{DBLP:conf/iccv/ZhuKZSUTF15} and English Wikipedia.
For unsupervised/supervised adaptive training on MDC and WOZ and fine-tuning on the GSD-overall dataset, all hyper-parameters are tuned on the GSD-overall validation set. For the first set of experiments on ATIS and Snips, the maximum sequence length is 50, the batch size is 128, and the number of training epochs is 30.  For the second set of experiments on multi-turn dialogues, the maximum sequence length is 128 and the batch size is 32. Adam~\cite{DBLP:journals/corr/KingmaB14} is used for optimization. The initial learning rate is 5e-5 for the supervised adaptive training and fine-tuning (in both sets of experiments), and 2e-5 for the unsupervised adaptive training. The dropout probability is 0.1.
The mask probability of the MLM task is 15\% for the unsupervised adaptive training. We compare using different numbers of previous user and system turns in the dialogue session and observe the best SLU performance from using all previous turns. 
The threshold $t_u$ for user act classification is selected from $[0.3, 0.4, 0.5]$ and tuned on the validation set.

\subsection{Results and Discussion}
\label{sec:results}
\subsubsection{Single-Turn SLU}
\label{subsubsec:single-turn-slu}

\begin{table*}[htb]
\begin{center}
\scalebox{0.9}{
\begin{tabular}{l |c c c |c c c}
\hline
\multirow{2}{*}{\textbf{Models}} & \multicolumn{3}{c|}{\textbf{Snips}} & \multicolumn{3}{c}{\textbf{ATIS}}  \\
 & \textbf{Intent} & \textbf{Slot} & \textbf{Frame} & \textbf{Intent} & \textbf{Slot} & \textbf{Frame} \\
 & \textbf{(Acc)} & \textbf{(F1)}  & \textbf{(Acc)} &  \textbf{(Acc)} & \textbf{(F1)} & \textbf{(Acc)} \\
\hline
RNN-LSTM~\cite{DBLP:conf/interspeech/Hakkani-TurTCCG16} & 96.9 & 87.3 & 73.2 & 92.6 & 94.3 & 80.7 \\
Atten.-BiRNN~\cite{DBLP:conf/interspeech/LiuL16} & 96.7 & 87.8 & 74.1 & 91.1 & 94.2 & 78.9 \\
Slot-Gated~\cite{DBLP:conf/naacl/GooGHHCHC18} & 97.0 & 88.8 & 75.5 & 94.1 & 95.2 & 82.6 \\
Capsule Neural Networks~\cite{DBLP:journals/corr/abs-1812-09471} & 97.3 & 91.8 & 80.9 & 95.0 & 95.2 & 83.4\\
\hline
(1) CELT w/o BERT pre-train & 97.8$\pm$0.2 & 90.0$\pm$0.6 & 79.3$\pm$1.4 & 96.9$\pm$0.1 & 92.7$\pm$0.1 & 80.5$\pm$0.4\\
(2) CELT w/o BERT pre-train + CRF & 97.9$\pm$0.3 & 90.8$\pm$0.2 & 80.9$\pm$0.5 & 97.0$\pm$0.3 & 93.1$\pm$0.2 & 81.6$\pm$0.4  \\
(3) (1) w/ BERT pre-train & \textbf{98.3}$\pm$0.3 & 96.4$\pm$0.2 & \textbf{91.9}$\pm$0.2 & {97.4}$\pm$0.4 & \textbf{95.9}$\pm$0.1 & \textbf{87.9}$\pm$0.4 \\
(4) (2) w/ BERT pre-train & \textbf{98.3}$\pm$0.1 & \textbf{96.5}$\pm$0.2 & 91.8$\pm$0.5 & \textbf{97.6}$\pm$0.1 & {95.7}$\pm$0.1 & 87.6$\pm$0.2\\
\hline
\end{tabular}
}
\end{center}
\caption{SLU performance on the single-turn Snips and ATIS testsets. Note that since Snips and ATIS are single-turn dialogues, all models in this table do not use context information. All models are trained and tested on the same training and test partitions of Snips and ATIS, respectively (no transfer learning is applied). The mean and standard deviation of SLU results from CELT w/o and with BERT pre-train, w/o and with replacing the softmax layer with a CRF layer, from 5 different models with different random initialization are given here. The metrics are the intent classification accuracy, slot filling F1, and sentence-level semantic frame accuracy. The results for the first group of models are cited from~\cite{DBLP:conf/naacl/GooGHHCHC18,DBLP:journals/corr/abs-1812-09471}.} 
\label{tab:single-turn:result}
\end{table*}

Table~\ref{tab:single-turn:result} shows the SLU performance as SF F1, IC accuracy, and sentence-level semantic frame accuracy on the Snips and ATIS datasets. The first group of models is considered as the baselines and it consists of the state-of-the-art joint IC and SF models: the sequence-based joint model using BiLSTM~\cite{DBLP:conf/interspeech/Hakkani-TurTCCG16}, the attention-based model~\cite{DBLP:conf/interspeech/LiuL16}, the slot-gated model~\cite{DBLP:conf/naacl/GooGHHCHC18}, and the capsule neural network based model~\cite{DBLP:journals/corr/abs-1812-09471}.

The second group of models in Table~\ref{tab:single-turn:result} includes the proposed CELT models. CELT with BERT pre-train significantly outperforms the baselines on both datasets. Compared to ATIS, Snips includes multiple domains and has a larger vocabulary. For the more complex Snips dataset, CELT with BERT pre-train achieves intent accuracy of 98.3\% (from 97.3\%), slot F1 of 96.4\% (from 91.8\%), and sentence accuracy of 91.9\% (from 80.9\%). On ATIS, CELT achieves intent accuracy of 97.4\% (from 95.0\%), slot F1 of 95.9\% (from 95.2\%), and sentence accuracy of 87.9\% (from 83.4\%). The gain from CELT with BERT pre-train on Snips over the baselines is much more significant on slot F1 and sentence frame accuracy than intent accuracy. Further analysis shows that 53.4\% of slots in the Snips test set can be found in Wikipedia. Since BERT pre-training data includes English Wikipedia, the model may have encoded the knowledge in representations and hence improves slot F1 and sentence accuracy. 

Without BERT pre-train, the SLU performance degrades drastically on both datasets. These results demonstrate the strong generalization and semantic representation capability of the BERT pre-train model, considering that it is pre-trained on large-scale text from mismatched domains and genres (books and Wikipedia). Without BERT pre-train, replacing the softmax layer with CRF consistently improves the sentence accuracy (3\% and 2.4\% relative gains for Snips and ATIS, respectively); whereas, adding CRF for CELT with BERT pre-train performs comparably. Hence, the second set of experiments uses CELT without CRF. 

Ablation analysis on Snips shows that when fine-tuning the BERT pre-train model separately for IC and SF, intent accuracy drops to $97.8 \pm 0.4$\% (from $98.3 \pm 0.3$\%), and slot F1 drops to $96.3 \pm 0.1$\% (from $96.4 \pm 0.2$\%). These results demonstrate that joint modeling in CELT improves the performance for both tasks. We compare CELT models with different fine-tuning epochs. The CELT model fine-tuned with only 1 epoch already outperforms the baselines in Table~\ref{tab:single-turn:result}. 

\begin{table*}[ht]
\begin{center}
\scalebox{0.9}{
\begin{tabular}{l | c c c c | c c c c | c c c c}
\hline
\multirow{3}{*}{\textbf{Models}} & \multicolumn{4}{c|}{\textbf{GSD-Resturant}} & \multicolumn{4}{c|}{\textbf{GSD-Movie}} & \multicolumn{4}{c}{\textbf{GSD-Overall}} \\
 & \textbf{Intent} & \textbf{Act} & \textbf{Slot} & \textbf{Frame} & \textbf{Intent} & \textbf{Act} & \textbf{Slot} & \textbf{Frame} & \textbf{Intent} & \textbf{Act} & \textbf{Slot} & \textbf{Frame} \\
 & \textbf{(Acc)} & \textbf{(F1)} & \textbf{(F1)} & \textbf{(Acc)} &  \textbf{(Acc)} & \textbf{(F1)} & \textbf{(F1)} & \textbf{(Acc)} &  \textbf{(Acc)} & \textbf{(F1)} & \textbf{(F1)} & \textbf{(Acc)} \\
\hline
RNN-NoContext~\cite{DBLP:conf/interspeech/GuptaRH18} & 83.61 & 87.13 & 94.24 & 65.51 & 88.51 & 93.49 & 86.91 & 62.17 & 84.76 & 89.03 & 92.01 & 64.56 \\
RNN-PreviousTurn~\cite{DBLP:conf/interspeech/GuptaRH18} & 99.37 & 90.10 & 94.96 & 86.93 & 99.12 & 93.58 & 88.63 & 77.27 & 99.31 & 91.13 & 93.06 & 84.19 \\ 
MemNet-20~\cite{DBLP:conf/interspeech/ChenHTGD16} & 99.67 & 95.67 & 94.28 & 89.52 & 98.76 & 96.25 & 90.70 & 80.35 & 99.29 & 95.85 & 93.21 & 86.92 \\
SDEN-20~\cite{DBLP:conf/sigdial/BapnaTHH17} & 99.84 & 94.43 & 94.81 & 89.46 & 99.60 & 97.56 & 90.93 & 82.55 & 99.81 & 95.38 & 93.65 & 87.50 \\
HRNN-SystemAct~\cite{DBLP:conf/interspeech/GuptaRH18} & \textbf{99.98} & 95.42 & 95.38 & 89.26 & 99.71 & 96.35 & 91.58 & 83.36 & \textbf{99.92} & 95.70 & 94.22 & 87.58 \\
\hline
CELT & 99.88 &	\textbf{98.47} &	\textbf{97.12} &	\textbf{95.63} &	\textbf{99.71} &	\textbf{98.45} &	\textbf{95.74} & \textbf{93.48} &	99.83 &	\textbf{98.47} &	\textbf{96.70} &	\textbf{95.02} \\
\hline
\end{tabular}
}
\end{center}
\caption{SLU performance on different test sets of the multi-turn GSD dialogue corpus, from baselines and our proposed CELT model, when trained on the GSD-overall training set. The results for the first group of models are cited from~\cite{DBLP:conf/interspeech/GuptaRH18}. MemNet-20 and SDEN-20 denote models with memory size 20.}
\label{tab:result}
\end{table*}

\begin{table}[ht]
\renewcommand{\arraystretch}{0.9}
\begin{center}
\scalebox{0.9}{
\begin{tabular}{l | c c c c}
\hline
\multirow{2}{*}{\textbf{Model}} & \textbf{Intent} & \textbf{Act} & \textbf{Slot} & \textbf{Frame} \\
 & \textbf{(Acc)} & \textbf{(F1)} & \textbf{(F1)} & \textbf{(Acc)} \\
\hline
a. CELT & 99.83 & 98.47 & 96.70 & \textbf{95.02} \\ \hline 
b. a-UA-SA & 99.62 & 98.29 &	94.99 &	93.00  \\ \hline \hline 
c. b+UA(ID) & 99.88  & 98.03 & 96.64 & 94.40  \\
d. b+UA(OOD) & 99.92 & 98.18 & 95.07 &93.35 \\
e. b+UA(ID)+SA(ID) & 99.90 & 98.43 & 96.17 & 94.60 \\  
f. b+UA(ID)+SA(ID+OOD) &99.83 & 98.47 & 96.70 & \textbf{95.02} \\
g. b+UA(ID+OOD) & 99.90 & 98.24 & 95.88  & 94.25  \\
h. b+UA(ID+OOD) &\multirow{2}{*}{99.92} &\multirow{2}{*}{96.58} &\multirow{2}{*}{96.58} & \multirow{2}{*}{94.81} \\ 
~~~~+SA(ID+OOD) & & & & \\
i. b+UA(ID+OOD+WOZ) & 99.92 & 98.52 &	95.02 &	93.71   \\
\hline \hline 
j. b-BERT pre-train & 99.92 &	92.44 &	91.12 &	86.54 \\
k. j-speaker embeddings & 99.96 &	92.22 &	90.43 &	86.15 \\
l. k-context utterances & 94.52 &	93.55 &	91.61 &	82.35  \\
m. l-system act embeddings & 77.33 &	89.11 &	90.93 &	66.88 \\
\hline
\end{tabular}
}
\end{center}
\caption{Ablation Analysis on the GSD-overall test set. UA and SA denote unsupervised and supervised adaptive training, respectively. ID and OOD denote in-domain and out-of-domain dialogues in MDC w.r.t. the GSD-overall test set, i.e., MDC restaurant and movie domain dialogues are ID data, MDC taxi domain dialogues are OOD data.  +WOZ denotes using MDC+WOZ data for adaptive training.}
\label{tab:ablation}
\end{table}

\subsubsection{Multi-Turn SLU and Transfer Learning}
\label{subsec:multi-turn-slu}
Table~\ref{tab:result} shows the IC accuracy, UAC F1, SF F1, and sentence frame accuracy on the GSD test sets. 
The first group of models includes the baselines.  \textit{RNN-NoContext}~\cite{DBLP:conf/interspeech/GuptaRH18} uses two-layer stacked BiRNN with GRU and LSTM cells respectively, and no context information is used.  \textit{RNN-PreviousTurn}~\cite{DBLP:conf/interspeech/GuptaRH18} is similar to the \textit{RNN-NoContext} model, with a different BiGRU layer encoding the previous system turn for slot tagging.  
\textit{MemNet}~\cite{DBLP:conf/interspeech/ChenHTGD16} uses memory network to encode the dialogue history utterances from both user and system. \textit{SDEN}~\cite{DBLP:conf/sigdial/BapnaTHH17} uses the dialogue history utterances from both user and system through a BiGRU for combining memory embeddings. \textit{HRNN-SystemAct}~\cite{DBLP:conf/interspeech/GuptaRH18} is the previous state-of-the-art (SOTA) system, using a hierarchical RNN to encode the dialogues acts of the previous system turn as the context information.

The second group of models includes the proposed CELT model after applying the proposed TL approaches. CELT achieves new SOTA results and the absolute gains from CELT over the previous SOTA user act F1 and sentence frame accuracy are 2.8\% and 6.11\% on the GSD-Restaurant testset, 0.89\% and 10.12\% on the GSD-Movie testset, and 2.62\% and 7.44\% on the GSD-Overall testset.
Table~\ref{tab:ablation} shows the ablation analysis on the GSD-Overall test set. Removing all unsupervised and supervised adaptive training for CELT (CELT-UA-SA) degrades the sentence accuracy from 95.02\% to 93.00\%. Further removing the unsupervised BERT pre-train step degrades sentence accuracy to 86.54\%. Particularly, user act F1 decreases from 98.29\% to 92.44\%, and slot F1 decreases from 94.99\% to 91.12\%. These results demonstrate that the contextual representations learned from the large-scale general purpose unlabeled text significantly help improve user act classification and  slot filling. After further removing the speaker embeddings, slot F1 drops from 91.12\% to 90.43\% and sentence accuracy drops from 86.54\% to 86.15\%, suggesting that CELT is capable of exploiting the additional discriminative information provided by the speaker embeddings. After further removing the context utterances, intent accuracy drops from 99.96\% to 94.52\% and sentence accuracy drops from 86.15\% to 82.35\%, indicating that the context utterances play a key role in intent prediction. After further removing system act embeddings, that is, no context is used, intent accuracy drops from 94.52\% to 77.33\%, user act F1 drops from 93.55\% to 89.11\% and sentence frame accuracy drops from 82.35\% to 66.88\%. These results show that using no context information degrades the SLU performance significantly. It is noticeable that although intent accuracy and slot F1 of CELT-NoContext (the last row in Table~\ref{tab:ablation}) are both lower than those of RNN-NoContext (the first row in Table~\ref{tab:result}), CELT-NoContxt achieves a better sentence accuracy (66.88\%) than RNN-NoContext (64.56\%), demonstrating the strength of CELT to enforce intent and slot coherence.

We further analyze the efficacy of using in-domain (ID) and out-of-domain (OOD) data for unsupervised adaptive (UA) and supervised adaptive (SA) training. As shown in Table~\ref{tab:ablation}, using OOD data for UA (model d) achieves small SLU improvement over model b. UA(ID) (model c) yields a significantly larger gain over model b compared to model d.
However, adding OOD data to ID data for UA (model g) degrades the performance slightly compared to model c. Adding WOZ to MDC ID+OOD data for UA (model i) further degrades the performance over model g. In contrast, after applying UA(ID) (model c), adding OOD to ID for SA (model f) outperforms SA(ID) (model e), achieving 95.02\% sentence frame accuracy.  These results suggest that SA can benefit from both ID and OOD data, probably due to the combined loss of IC and SF. We will explore different losses for UA other than MLM and NSP losses in order to benefit from both ID and OOD data.

We also observe fast convergence speed of both CELT-UA-SA and CELT models on the GSD-overall testset, consistent with previous observations on models using BERT pre-train. For CELT-UA-SA, the user act F1 and sentence frame accuracy increase from 87.50 and 52.52 (epoch 1) to 97.19 and 90.31 (epoch 5), and keep improving until epoch 20 (98.29 and 93.00) but degrade from epoch 20 to 40. For CELT, these two results increase from 92.34 and 62.60 (epoch 1) to 97.17 and 93.79 (epoch 5), and keep improving from epoch 5 to 40 (98.47 and 95.02). 

\section{Conclusions}
We propose Context Encoding Language Transformer for SLU facilitating exploiting various context information and different transfer learning approaches for leveraging external resources. Experimental results demonstrate that CELT with TL achieves new SOTA SLU performance on two large-scale single-turn dialogue benchmarks and one multi-turn dialogue benchmark. Future work includes improving supervised and unsupervised TL and exploring TL on knowledge bases.

% References should be produced using the bibtex program from suitable
% BiBTeX files (here: strings, refs, manuals). The IEEEbib.bst bibliography
% style file from IEEE produces unsorted bibliography list.
% -------------------------------------------------------------------------
\bibliographystyle{IEEEbib}
\bibliography{mybib}

\end{document}